\documentclass[10pt,twocolumn,letterpaper]{article}

\usepackage{wacv}              
\usepackage{graphicx}
\usepackage{amsmath}
\usepackage{amssymb}
\usepackage{booktabs}

\usepackage[pagebackref,breaklinks,colorlinks]{hyperref}

\usepackage[capitalize]{cleveref}
\crefname{section}{Sec.}{Secs.}
\Crefname{section}{Section}{Sections}
\Crefname{table}{Table}{Tables}
\crefname{table}{Tab.}{Tabs.}

\def\wacvPaperID{*****} 
\def\confName{WACV}
\def\confYear{2025}

\begin{document}

\title{\textit{Similarity Trajectories}: Linking Sampling Process to Artifacts in Diffusion-Generated Images}
\vspace{-1em}
 \author{Dennis Menn \qquad Feng Liang \qquad Hung-Yueh Chiang \qquad Diana Marculescu \\[0.2em]
Chandra Family Department of Electrical and Computer Engineering\\[0.2em]
The University of Texas at Austin\\
{\tt\small \texttt{\{dennismenn,jeffliang,hungyueh.chiang,dianam\}@utexas.edu}}
}
\maketitle

\begin{abstract}
   Artifact detection algorithms are crucial to correcting the output generated by diffusion models. However, because of the variety of artifact forms, existing methods require substantial annotated data for training. This requirement limits their scalability and efficiency, which restricts their wide application. This paper shows that the similarity of denoised images between consecutive time steps during the sampling process is related to the severity of artifacts in images generated by diffusion models. Building on this observation, we introduce the concept of \textit{Similarity Trajectory} to characterize the sampling process and its correlation with the image artifacts presented. Using an annotated data set of 680 images, which is only 0.1\% of the amount of data used in the prior work, we trained a classifier on these trajectories to predict the presence of artifacts in images. By performing 10-fold validation testing on the balanced annotated data set, the classifier can achieve an accuracy of 72.35\%, highlighting the connection between the \textit{Similarity Trajectory} and the occurrence of artifacts. This approach enables differentiation between artifact-exhibiting and natural-looking images using limited training data.
\end{abstract}

\begin{figure*}[!h]
\centering
\vspace{-1.6cm}
\includegraphics[width=0.75\linewidth]{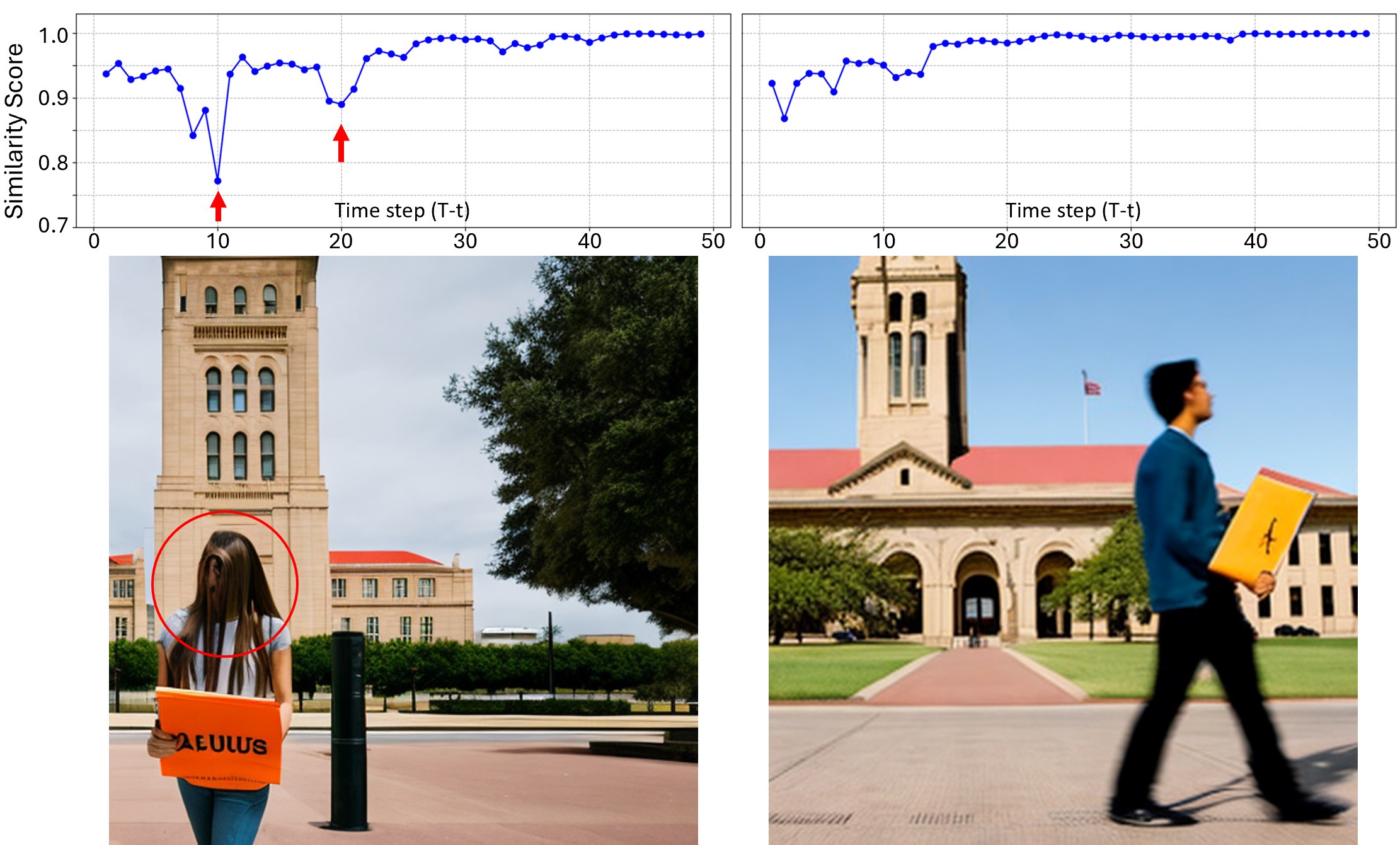} 
\caption{Comparison of images exhibiting strong artifacts (left) versus a more natural appearance, alongside their corresponding \textit{Similarity Trajectories}. The image on the left displays pronounced artifacts, particularly in the circled area where the subject's face blends unnaturally with their hair. This is reflected in the \textit{Similarity Trajectory}, which is more erratic and shows a significant drop, as indicated by the red arrows. In contrast, the right image appears more natural, with a smoother \textit{Similarity Trajectory} that exhibits consistency. The prompt for the images is "\textit{A student walking in front of the UT tower, with one hand holding a calculus book.}"}
\vspace{-0.3cm}
\label{pic:demo}
\end{figure*}

\section{Introduction}
\label{sec:intro}

Image generation models have recently garnered significant attention for their diverse applications in image inpainting, super-resolution, style transfer, image deblurring, etc. \cite{Rombach_2022_ldm, chung2023diffusion, Li2022SRDiff, Lugmayr_2022_CVPR, Kawar_2022_Denoising}. These technologies have had a profound impact across various industries, including advertising, entertainment, and design. Despite their ability to produce impressive results, these models often suffer from a critical drawback: the generation of images with noticeable artifacts. Common problems include distorted object shapes or merging of distinct objects \cite{Liang_2024_rich}.

The presence of artifacts in the generated images highlights the importance of detecting these imperfections. This is because after the artifacts are detected, the diffusion model can then inpaint the specific area, thereby improving the quality of the images generated \cite{xu2023imagereward, Liang_2024_rich}. Recognizing the importance of this task, recent research has focused on collecting large annotated data sets to predict human ratings \cite{neurips2023Kirstain, xu2023imagereward, wu2023hps}. A particularly noteworthy study by Liang \textit{et al}.\cite{Liang_2024_rich} focused on developing a data set with annotated heat maps that precisely identify where and how the image generation process fails. This line of work has proven valuable for artifact detection and provides explainable results. However, creating such a data set requires a vast amount of labeled data and meticulously curated annotations that pinpoint the specific area causing image generation to fail. This process requires considerable effort, which in turn limits the broader application of these methods.

In this paper, instead of focusing on the final generated image, we adopt a different approach that takes into consideration the sampling process. Specifically, we analyze the variation in similarity between consecutive denoised images throughout the sampling process. This analysis provides valuable insights into the severity of artifacts in the generated image. We conjecture that this is because each iteration in the sampling process progressively improves the denoised image by building upon the results of the previous step. If consecutive denoised images are dissimilar, this process is similar to adding two different images. This can lead to merged objects, eventually resulting in artifacts in the final image. Hence, we conjecture that, by monitoring the similarity between consecutive denoised images during the sampling process, we would be able to assess the severity of artifacts present in images.

To capture the behavior of the sampling process, we introduce the concept of a \textit{Similarity Trajectory}. This trajectory is defined as the similarity between denoised images in consecutive time steps throughout the entire sampling process. Therefore, a drop or low similarity between consecutive denoised images likely signals the presence of artifacts in the generated image. Figure \ref{pic:demo} illustrates this with an example: on the left, an image with pronounced artifacts, where the girl's face blends unnaturally with her hair, and on the right, a more natural-looking image. Both images are generated from the same prompt \textit{"A student walking in front of the UT tower, with one hand holding a calculus book"}. We emphasize that a significant advantage of detecting artifacts based on the \textit{Similarity Trajectory} is that it allows the use of a very small data set to train detection algorithms, as few as 680 training images, as the trajectory provides a much more condensed representation compared to raw images. In contrast, previous methods cannot achieve this efficiency because they rely on processing raw images, which requires over a million training images \cite{neurips2023Kirstain}.

The main contributions of this paper are summarized below:
\begin{itemize}
    \item We introduce the concept of \textit{Similarity Trajectory} to characterize the sampling process's behavior for diffusion models.
    \item Through extensive experimental results, including human evaluation testing, we uncover the correlation between the \textit{ Similarity Trajectory} and the presence of artifacts in generated images.
    \item We further demonstrate that the \textit{Similarity Trajectory} can be used to evaluate the model's performance in generating images, under the EDM2 model framework \cite{Karras2024edm2}.
\end{itemize}

This paper is organized as follows. In Section 2, we first review related work in artifact detection. We then explain in Section 3 how the \textit{Similarity Trajectory} is obtained and why it is linked to the presence of image artifacts. In Section 4, we detail the experimental configuration used to develop a classifier aimed at detecting image artifacts using the \textit{Similarity Trajectory}. Additionally, we validate the severity of artifacts in the chosen images through a human evaluation test. In Section 5, we present experimental results that support our claim that the \textit{Similarity Trajectory} is related to the presence of image artifacts.
\section{Related Work}
Over the years, image generation technology has undergone significant advancements \cite{kingma2022vae, NIPS2014GAN, neurips2020ddpm, Rombach_2022_ldm}. One of the first foundational approaches is the Variational Auto Encoder (VAE), which learns to map images to and from a latent space \cite{Rombach_2022_ldm}. This capability enables VAEs to generate images by randomly sampling the latent space.

In recent years, Diffusion Models have achieved state-of-the-art performance in generative tasks \cite{nips2021_dmbeatsgan}. In this framework, images are generated by multiple iterations of the denoising step \cite{neurips2020ddpm}. The iterative nature of the diffusion model enables it to produce high-quality data, but at the same time requires substantial computational demands \cite{neurips2020ddpm}. To address this, the concept of Latent Diffusion Models (LDM) was introduced \cite{Rombach_2022_ldm}. This approach, combined with Auto Encoders, generates images in a lower-dimension latent space, thereby significantly reducing computational costs. Consequently, this development has made the concept of large-scale text-to-image models feasible \cite{Rombach_2022_ldm, podell2024sdxl}.

Despite their astonishing performance, a significant portion of images generated by diffusion models still suffer from artifacts. To address this issue, recent research has focused on detecting these flawed images and fixing the artifacts \cite{wu2023hps, neurips2023Kirstain, Liang_2024_rich, xu2023imagereward}. A common approach involves labeling the generated images according to human preference and training a model to predict the human preference \cite{wu2023hps, xu2023imagereward, neurips2023Kirstain}. For example, the "Pick-a-Pic" approach creates a data set of more than 500,000 pairs of images, each pair consisting of two images with a human-preferred choice \cite{neurips2023Kirstain}. This data set is then used to fine-tune a CLIP-based model to predict human preferences \cite{pmlr-v139-radford21a}. Similarly, "ImageReward" collects a data set by ranking and rating images based on their quality. This data set is then used to train a model, which is subsequently used to fine-tune the diffusion model, enhancing the quality of generated images\cite{xu2023imagereward}.

Another approach involves providing more detailed information to the model beyond just human preference scores. This is done by annotating specific areas of the image where artifacts occur \cite{Zhang_2023_ICCV, Liang_2024_rich}. In Liang \textit{et al.} \cite{Liang_2024_rich}, the author extends this method by labeling not only artifact regions but also areas with misaligned keywords. This additional information helps the model learn more effectively by offering a deeper understanding of the artifacts.

Different from previous studies that judge the quality based on the final generated images, our work uncovers the insights in the sampling process itself which provides fruitful information on whether an image contains artifacts. To our knowledge, this is the first work to reveal this relationship. A direct benefit of uncovering this relationship is that the training data required for classification is reduced by orders of magnitude compared to previous methods.

\section{Background Knowledge}
To analyze the dynamics of the time-series data (\textit{Similarity Trajectory}), we rely on the Haar Transform. This transformation allows us to identify significant fluctuations and trends within the data. The subsequent section delves into the details of the Haar Transform.

\paragraph*{Haar Transformation}
The Haar Transform decomposes the original time series data \( \{x_t\}_{t=1}^{T} \) into approximation and detail coefficients at various scales, capturing both global trends and local variations.

At the first level of decomposition, for $k = 1, 2, \dots, \left\lfloor \dfrac{T}{2} \right\rfloor$, the approximation coefficients $a_1(k)$ and detail coefficients $d_1(k)$ are calculated as:
\begin{equation}
a_1(k) = \frac{ x_{2k} + x_{2k+1} }{ 2},
\end{equation}
\begin{equation}
d_1(k) = \frac{ x_{2k} - x_{2k+1} }{ 2 }.
\end{equation}

This process is recursively applied to the approximation coefficients to obtain higher-level coefficients. At level $j+1$, the coefficients are computed as:
\begin{equation}
a_{j+1}(k) = \frac{ a_j(2k) + a_j(2k+1) }{ 2 },
\end{equation}
\begin{equation}
d_{j+1}(k) = \frac{ a_j(2k) - a_j(2k+1) }{ 2 },
\end{equation}
where $j = 1, 2, \dots, J$, and $J$ is the maximum level of decomposition. From the transformation, we obtain a set of approximation coefficients $\{ a_j(k) \}$ as well as detail coefficients $\{ d_j(k) \}$ corresponding to each basis function at various levels. Note that the detail coefficients capture the fluctuations in the time-series data, which are important for analyzing the proposed \textit{Similarity Trajectory} and assessing the presence of artifacts in images.

\section{Similarity Trajectory}
\label{sec:sim_traj}
At a high level, the \textit{Similarity Trajectory} represents the similarity between denoised images in consecutive time steps during the sampling process of a diffusion model. In the following, we present the mathematical definition of the \textit{Similarity Trajectory} and then explain the intuition behind its importance.

We define the \textit{Similarity Trajectory} as a time series \( \{z_t\}_{t=T-1}^{1} \), as follows:
\begin{equation}
\{\, z_t \,\}_{t=T-1}^{1}, \quad \\ \text{where} \  z_t = d\left( x_0^{(t)},\ x_0^{(t-1)} \right).
\end{equation}

\( z_t \) represents the similarity score between the denoised images \( x_0 \) predicted in consecutive time steps \( t \) and \( t-1 \). The time index \( t \) adheres to the convention used in previous research, indicating the sequential steps of the forward diffusion process. The similarity metric \( d(\cdot, \cdot) \) quantifies the similarity of two images: the more similar the images, the higher the value of \( d \). 



The \textit{Similarity Trajectory} can be computed for different diffusion model training frameworks, provided that the model is trained to predict denoised images from noise corruption. This is because the \textit{Similarity Trajectory} essentially consists of a sequence of similarity scores between denoised images in consecutive time steps. Therefore, we can calculate the \textit{Similarity Trajectory} as long as the denoised images can be derived from the diffusion model. In the following, we provide the mathematical formulas for \( x_0^{(t)} \) in the context of Stable Diffusion 2 (SD2) using the DDIM sampler and the EDM2 framework \cite{Karras2024edm2}.

\begin{figure*}[!h]
\centering
\vspace{-1.3cm}
\includegraphics[width=0.8\linewidth]{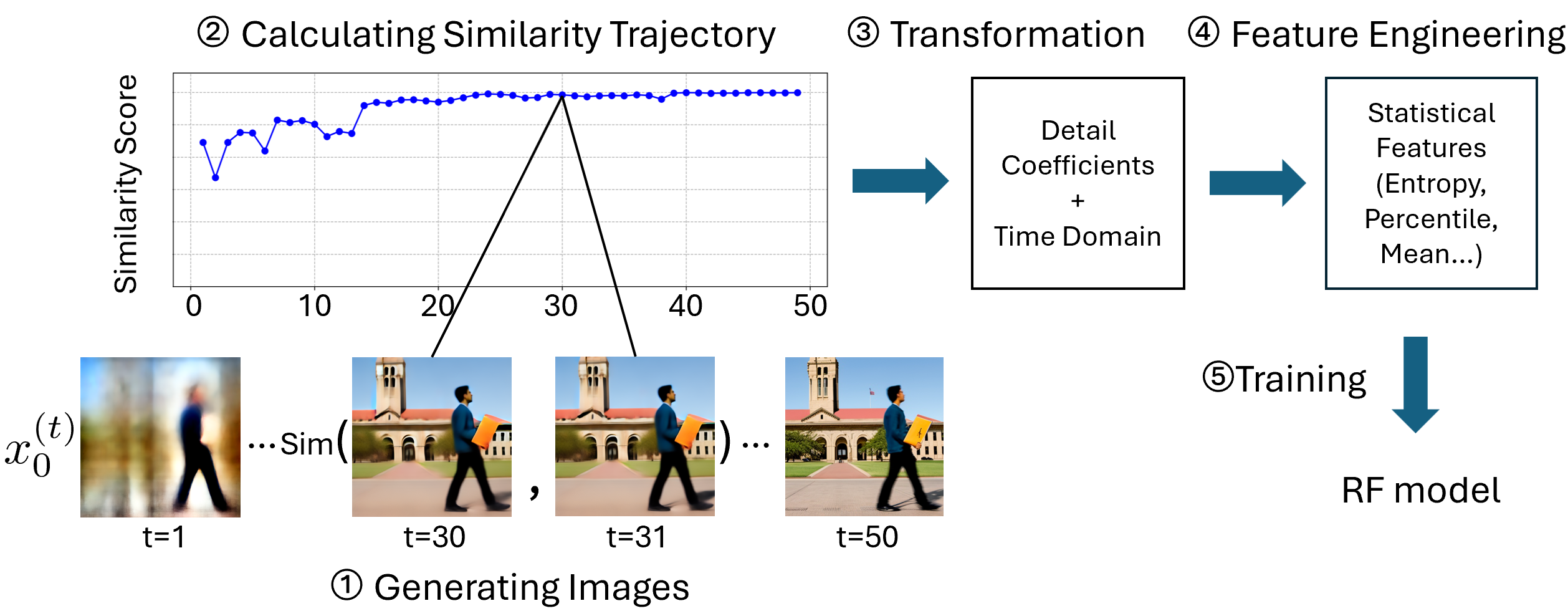} 
\caption{Flowchart illustrating the methodology for training a Random Forest (RF) classifier to detect artifacts in images based on the \textit{Similarity Trajectory}. The process involves: (1) generating images and recording the denoised images \(x_0^{(t)}\) at each time step; (2) calculating the similarity between consecutive denoised images \(x_0^{(t)}\) and \(x_0^{(t+1)}\) to construct the \textit{Similarity Trajectory}; (3) applying Haar transform to the \textit{Similarity Trajectory} to obtain sets of detailed coefficients and dividing the original \textit{Similarity Trajectory} into time-domain trajectory sets; (4) performing feature engineering by extracting statistical properties from each set; and (5) using the extracted features to train the RF classifier for classifying the presence of artifacts in the generated images.} 
\vspace{-0.4cm}
\label{pic:flow_chart}
\end{figure*}

\paragraph{Denoised Image from SD2 with DDIM Sampler.}

Equation \ref{eqn:ddim} illustrates the method for computing $x_0^{(t)}$, which represents the denoised image at the specific time step $t$.

\begin{equation}
x_0^{(t-1)} = \frac{x_t - \sqrt{1 - \alpha_t} \cdot \epsilon_\theta^{(t)} (x_t)}{\sqrt{\alpha_t}}
\label{eqn:ddim}
\end{equation}

We follow the notation from prior work \cite{song2021denoising} for Equation \ref{eqn:ddim}. \( x_t \) represents the noisy image, \textit{i.e.,} input to the diffusion model. \( \epsilon_{\theta}^{(t)}(\cdot) \) denotes the noise predicted from the U-net in SD 2 at time step \( t \). \( \alpha_t \) represents the cumulative product \(\prod_{i=1}^{t}(1-\beta_i)\). The term $\beta_i$ is defined as $\frac{i}{T}$, where $T$ indicates the total number of steps of denoising.

\paragraph{Denoised Image from EDM2 with Heun Sampler.}
In Equation \ref{eqn:heun}, we show how to calculate the denoised images for the Heun Sampler \cite{karras2022elucidating}.

\begin{equation}
\label{eqn:heun}
x_0^{(t-1)} = x_t - \frac{1}{2} \sigma_t \cdot (n_t+n_t')
\end{equation}

In Equation \ref{eqn:heun}, \( x_0^{(t-1)} \) represents the denoised image at time step \( t-1 \), and \( x_t \) is the current noisy image at time step \( t \). \( \sigma_t \) is the predetermined noise level at the time step $t$. The terms \( n_t \) and \( n_t' \) are the predicted noise, normalized to a standard deviation of 1, during the first and second orders of the Heun method, respectively.

\section{Methodology}
\label{sec:artifact_process}
Through extensive research, we observe that an apparent and sustained decline in the \textit{Similarity Trajectory} suggests the presence of artifacts in the generated image. Given a similarity trajectory \(\{ z_t \}_{t=T-1}^{1}\), the maximum decline \(D_{\text{max}}\) is defined as:
\[
D_{\text{max}} = \max_{\substack{T-1 \leq s < e \leq 1 \\ z_s > z_{s+1} > \dotsb > z_e}} \left( z_s - z_e \right).
\]

This represents the greatest decrease in similarity scores over any continuous strictly decreasing subsequence within the trajectory. We then independently calculate the mean of these maximum decline values for both artifact-exhibiting and natural-looking images.

This decline suggests a misalignment between denoised images in consecutive time steps. We believe that such a misalignment can lead to artifacts because, during the sampling process, each iteration involves subtracting a small amount of predicted noise from the noisy image (latent). This process effectively removes a portion of the noisy image (latent) and replaces it with the denoised image (latent). As the diffusion model determines the contours of the objects, this blending can merge objects predicted at different time steps, resulting in distortions of their original shapes. This may lead to artifacts in the generated images.

\begin{figure}[t]
\centering
\includegraphics[width=1.\columnwidth]{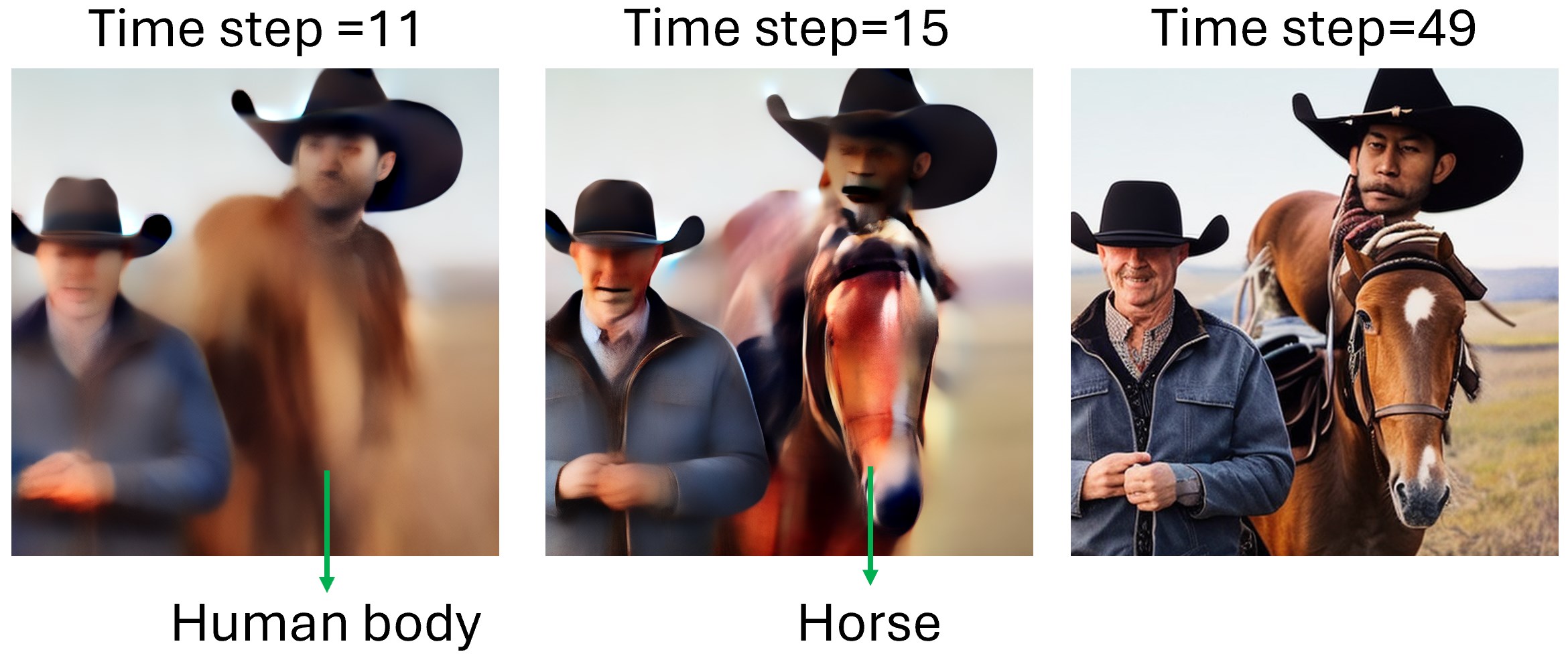} 
\caption{Artifact formation in the sampling process. The denoised images $x_0$ at various time steps illustrate that changes in the diffusion model's predictions between consecutive steps can cause overlapping objects. This overlap may distort the original shapes, leading to the presence of artifacts. The prompt for the image is "A man in a jacket and cowboy hat and a person on a horse".}
\label{pic:artifact_forming}
\vspace{-.4cm}
\end{figure}

Figure~\ref{pic:artifact_forming} illustrates an example of how artifacts can form. The images shown are the denoised images \(x_0\) at various time steps \(T - t\). At time step 11, two men are standing together. However, in the subsequent time step, a different prediction leads to the overlapping of distinct objects, such as a horse and one of the men's bodies, which eventually results in artifacts in the final time step. 
It is important to note that artifacts can originate from multiple sources. Apart from the merging of different objects, artifacts may also occur when the prediction of the same object changes shape across consecutive time steps. Moreover, the formation of artifacts is frequently unclear and elusive, likely because the denoising occurs within the latent space instead of the direct image space, coupled with the artifacts arising at earlier time steps in the sampling process.

A straightforward method to determine if there is a difference in the \textit{Similarity Trajectory} between artifacts-exhibiting and natural-looking images involves compiling a data set of generated images classified as such. We first investigate whether artifact-exhibiting images have more significant drops in their \textit{Similarity Trajectories} by calculating the maximum drop in these trajectories. We then use the data set to train a classifier, aiming to predict the presence of artifacts based on \textit{Similarity Trajectories}. If the classifier can identify artifacts in unseen images based on the \textit{Similarity Trajectory}, it will provide evidence that the behavior of the \textit{Similarity Trajectory} is related to image artifacts.

\subsection{Calculating Similarity Trajectory}
Given a similarity metric \( d(\cdot, \cdot) \), we calculate the trajectory for each image as \( \{z_t\}_{t=T-1}^{1} \), where \( z_t = d(x_0^{(t)}, x_0^{(t-1)}) \). Here, \( x_0^{(t)} \) represents the denoised latents from the U-net \cite{unet} of SD2, which are then decoded into pixel space via the VAE. The chosen similarity metric, DreamSim, is a neural network-based method that operates in the pixel space and provides similarity scores that align with human perception \cite{fu2023dreamsim}. We selected DreamSim because humans are sensitive to abrupt changes in denoised images, making a human-aligned metric beneficial for this task. To ensure that the DreamSim metric increases with the similarity between two images, we define the similarity metric as one minus the output score of DreamSim.
\subsection{Training Classifier}
\label{sec:transform}
We select a Random Forest (RF) Classifier \cite{breiman2001random} as our classifier due to its strong performance, especially when the amount of training data is scarce. Before inputting the \textit{Similarity Trajectory} into the model, we apply feature engineering to preprocess the \textit{Similarity Trajectory}. 

Our goal is to detect drops in the \textit{Similarity Trajectory}. To achieve this, we apply Haar transformation to the entire trajectory because of its ability to detect sudden changes in time series data \cite{stankovics200325}. For each basis in the Haar transformation, we will obtain a corresponding set of detail coefficients. Additionally, we divide the entire \textit{Similarity Trajectory} into three equal sets based on time steps, as well as considering the entire trajectory as a single set.

For all the sets obtained, whether they are detail coefficients from the Haar Transform or in the time domain, we calculate ten statistical features for each set. They include entropy, 5th percentile, 25th percentile, median, 75th percentile, 95th percentile, mean, standard deviation, mean crossings, and zero crossings. We calculate percentile and mean because in the detail coefficients, these quantities indicate fluctuations within the Similarity Trajectory. Standard deviation and entropy can characterize the fluctuations in the time domain. Mean crossings ($C_{\mu,S}$) quantifies how frequently the \textit{Similarity Trajectory} oscillates around its mean value, providing insight into the rapidity of these changes. Similarity, zero crossings ($C_{0,S}$) counts how often the data crosses zero, which, in the context of detail coefficients, reflects the number of times the \textit{Similarity Trajectory} changes direction from monotonically increasing to decreasing or vice versa. The mathematical definitions for the number of mean crossings and zero crossings are provided below.

 \noindent \textbf{Mean Crossings ($C_{\mu,S}$)}: 
\begin{equation}
C_{\mu,S} = \sum_{i = 1}^{N_S} \mathbb{I} \left[ \left( S_{i+1} - \mu_s \right) \left( S_i - \mu_s \right) < 0 \right],
\end{equation}
where $N_S$ is the total number of elements in set $S$. $S_i$ is the $i^{th}$ element in $S$ and $\mathbb{I}[\cdot]$ is the indicator function. $\mu_S$ indicates the mean of the set.\\

\noindent \textbf{Zero Crossings ($C_{0,S}$)}: 
\begin{equation}
C_{0,S} = \sum_{i = 1}^{N_S} \mathbb{I} \left[ S_{i+1} S_i < 0 \right].
\end{equation}
where $N_S$ is the total number of elements in set $S$. $S_i$ is the $i^{th}$ element in $S$ and $\mathbb{I}[\cdot]$ is the indicator function.

All of the described features serve as inputs to the RF Classifier. Additionally, we incorporate prediction probabilities obtained from a \textit{k}-Nearest Neighbor (\textit{k}-NN) model trained on the trajectory \cite{Cover67knn}. For further details, please refer to the Supplementary Material.

\subsection{Evaluating Models}
When evaluating the model, we also hypothesize that a better-performing generative model—characterized by its ability to produce higher-quality images—will generate fewer images displaying artifacts. Consequently, such a model is expected to exhibit fewer drops in \textit{Similarity Trajectories}, since these drops signal the presence of artifacts, resulting in a higher average \textit{Similarity Trajectory} compared to a weaker model.

\section{Experimental Setup}
\label{sec:Exp_setup}
In the following, we introduce the experimental setup to detect artifacts based on \textit{Similarity Trajectories}.

\subsection{Labeling Datasets}
\label{sec:dataset}
We randomly select 250 prompts from the 2014 MS COCO data set\cite{cocodataset}. Using SD2 with the DDIM 50 time step sampler \cite{Rombach_2022_ldm}, we generate nine images per prompt, resulting in a total of 2,250 images. To ensure the quality of our data set, we manually review these images, selecting only those with obvious artifacts or those that appear natural-looking (without or mostly without artifacts). Images that fall between these categories are discarded, resulting in 425 natural-looking and 255 artifact-exhibiting images, for a total of 680 images. We emphasize that the selection process focuses solely on the quality of the final generated image as opposed to looking at the \textit{Similarity Trajectory}.

\subsection{Experimental Settings for Model Evaluation}
Our objective is to examine whether there is a connection between the \textit{Similarity Trajectory} and the performance of the model, defined as being able to generate better quality images. Although the image generation model is typically evaluated through the FID score \cite{fid_Martin}, the FID score is known to be inaccurate in judging the model's performance and does not align with human perception \cite{podell2024sdxl}.

For our experiments, we use models provided in prior work \cite{Karras2024edm2}. These models are trained and evaluated within a consistent framework, ensuring that differences between them are limited to model size and the number of training steps. This allows us to assess model performance in a more controlled way, making it easier to determine whether such a relation exists. We conducted two types of comparison in the experiment: one controlling for model size and the other for training steps.

To investigate such a relationship, we first average the \textit{Similarity Trajectory} across 5,000 images. This is because we want to assess the overall performance of the model instead of individually generated images. In the experiment, we use the Heun sampler \cite{Karras2024edm2} with 32 inference steps and set the $S_{churn}$ parameter at 40. The guidance scale is configured to 1.5, which helps align the generated images more closely with the real data distribution when utilizing classifier-free guidance. In particular, the unconditional model for classifier-free guidance is selected to be identical to the conditional model but does not receive any prompt. This approach ensures that our evaluation is focused solely on the performance of the model being tested.

Next, we directly averaged the 5,000 \textit{Similarity Trajectories} without decoding them into pixel space. The similarity metric used is the Root Mean Square Error (RMSE) between adjacent time steps. We intentionally avoid decoding the latent representations into pixel space to eliminate the influence of decoder sensitivity. This is because we aim to establish a more robust metric for evaluating the diffusion model, independent of decoder-induced variations.

\section{Experimental Results}
To assess whether there is a difference between \textit{Similarity Trajectories} from artifact-exhibiting and natural-looking images, we perform experiments using the labeled data set. 

\subsection{Similarity Trajectory Maximum Decline}
\label{sec:std}
We analyze the maximum decline within individual trajectories, which follows directly from our observation that artifact-exhibiting images will experience a more severe decline of the \textit{Similarity Trajectory}. 

For natural-looking images, the average maximum decline is \(0.017 \pm 0.0011\), where 0.0011 represents the Standard Error of the Mean (SEM). In comparison, the artifacts-exhibiting images show an average drop of \(0.027 \pm 0.0015\). Note that the difference between the average maximum decline of natural-looking images and artifacts-exhibiting images far exceeds the SEM by a factor of 10$\times$, suggesting such a relation indeed exists. These results indicate that the drop for artifact-exhibiting images is indeed larger than that for natural-looking images, thereby providing statistical support for our observations.

\begin{figure}[t]
\centering
\includegraphics[width=1\columnwidth]{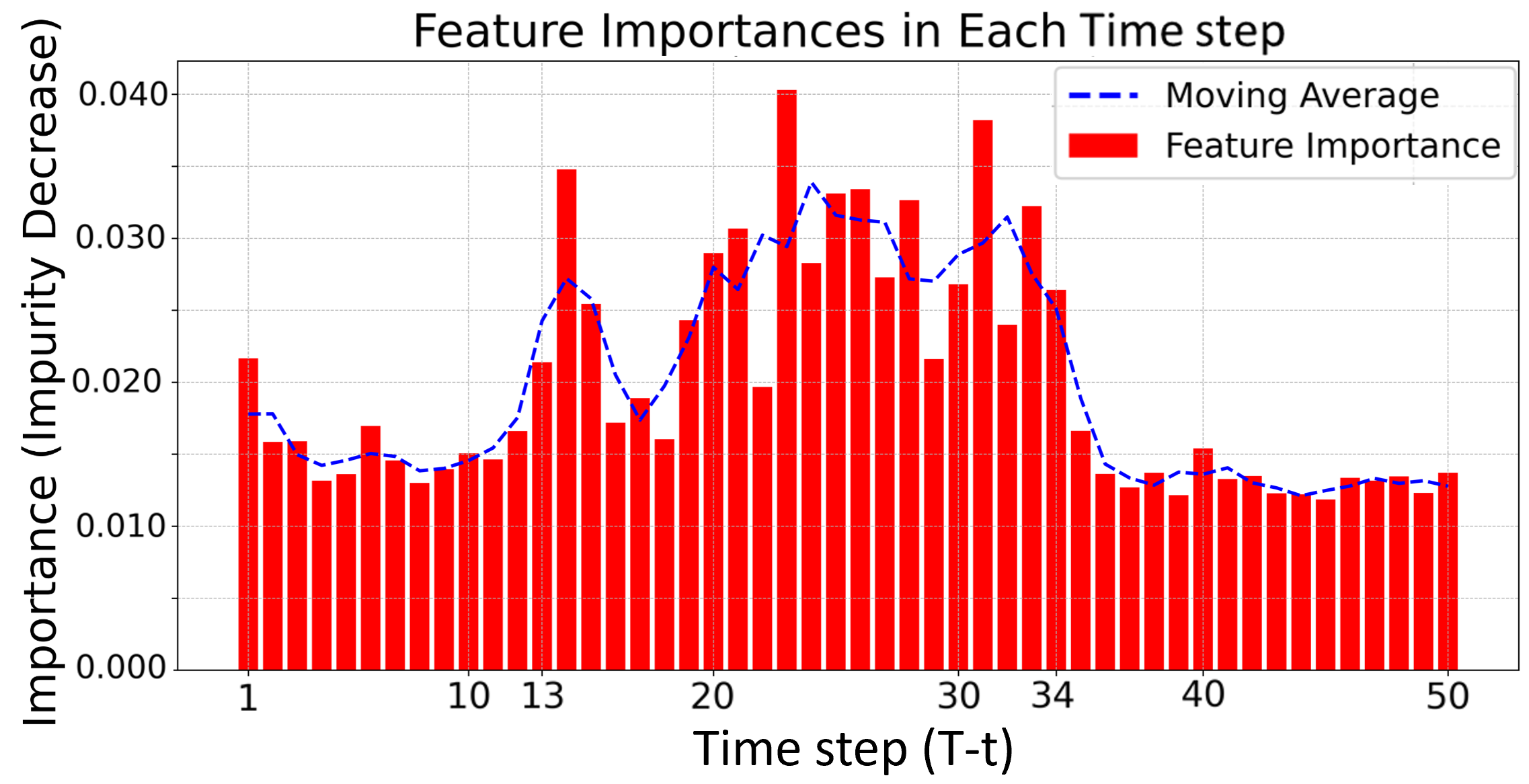} 
\caption{Average Gini impurity reduction at each time step from the RF Classifier. We input raw \textit{Similarity Trajectory}, without any transformations, into the RF classifier and analyze the average Gini impurity reduction.}
\label{pic:tree_importance}
\end{figure}

Note that in this experiment, we calculate the maximum decline only within the middle section of the time steps (\textit{i.e.,} time steps 13 to 34) of the \textit{Similarity Trajectory}. This is because prior work indicated that the middle section of the sampling process has the greatest influence on the final image \cite{zhang2023shiftddpms}. We also carried out experiments to find which part of the \textit{Similarity Trajectory} is critical for identifying artifacts. We feed the unprocessed trajectory directly into the RF Classifier and examine the average decrease in the Gini impurity of the RF. A decrease in Gini impurity indicates that specific parts of the trajectory are important for classification. The results are shown in Figure \ref{pic:tree_importance}. 
We observe an obvious surge of impurity decrease between time steps 13 and 34, indicating that this range is critical for classifying image quality. As a result, our analysis only focuses on this section. 

\subsection{Trajectory-Based Image Quality Classification} \label{sec:tree}
To explore whether the statistical differences between trajectories can serve as predictors of image quality, we use a RF Classifier of 1,000 trees and apply feature engineering to the trajectory mentioned in the Sec. \ref{sec:Exp_setup}. We then train the classifier on the balanced data set of 255 natural-looking images and 255 artifact-exhibiting images. Given this balanced set-up, random guessing would result in an accuracy of only 50\%. The 255 natural-looking images are selected as the first 255 images from that of the original training data set. 

By performing a 10-fold validation test, we obtained an accuracy of 72.35\% with a SEM of 2.1\%. This is significantly higher than random guessing, indicating a notable correlation between the \textit{Similarity Trajectory} and image quality. 

\subsection{Real-World Setting Evaluation}
To determine whether the RF-based method can classify images with and without artifacts in a real-world setting, we rely on the testing criteria from prior work \cite{neurips2023Kirstain}. 

We begin by randomly selecting 100 prompts from the Pick-a-Pic data set \cite{neurips2023Kirstain}. Using SD2, we generate 100 images per prompt, each with a different random seed. Among the 100 generated images, we use the RF Classifier to predict the probability that each image contains artifacts, using their respective trajectories. We select the images with the highest and lowest probability of having artifacts for each prompt and shuffle and pair them, resulting in 100 test pairs.

To evaluate how closely our classifier's predictions match human judgments on image artifacts, we recruited 10 human participants and divided the 100 test pairs into two groups of 50, with each group evaluated by five participants. For each test pair, participants compare the two images and select the one they believe exhibits more severe artifacts, defined as notable structural or shape deviations from real-world images. If participants cannot decide which image shows more severe artifacts, they can declare a draw to avoid random guessing. Our goal is to determine whether human judgments align with the classifier's selection on unseen data, thereby assessing the classifier's performance.

In the human evaluation test, we found that, on average, human participants' selections aligned with the classifier 58.1\% of the time, differed 21.7\% of the time, and resulted in a draw 20.2\% of the time, as shown in Table \ref{table:evaluation_results}. These results indicate that participants were approximately 2.7 $\times$ more likely to agree with the RF Classifier than to disagree, suggesting that the \textit{Similarity Trajectory} is indeed correlated with image artifacts.

\begin{table}[h]
\centering
\caption{Results of Human Evaluation Test. Human participants' selections matched that of the classifier in 58.1\% of cases, differed in 21.7\%, and could not decide 20.2\%.}
\label{table:evaluation_results}
\begin{tabular}{l c}
\hline
\textbf{Outcome} & \textbf{Percentage (\%)} \\
\hline
Selections Matched    & 58.1 \\
Selections Differed   & 21.7 \\
Draw                  & 20.2 \\
\hline
\end{tabular}
\end{table}

As a reference, prior work \cite{neurips2023Kirstain} evaluates the model by forming pairs composed of the best of 100 images evaluated by the model and a randomly selected image, rather than the worst. Then human participants decide which image is better in a pair but the draw is not allowed. On average, 71.4\% of human selection aligns with that of the model, highlighting the difficulty of the task. Whereas our classifier only utilizes less than 0.1\% of the training data, compared with the prior work, to train an artifact detection model, demonstrating the effectiveness of \textit{Similarity Trajectory} when sparse annotated training data is accessible.

\subsection{Averaging Trajectories for Model Assessment}
In this section, we present the results of experiments designed to explore the relationship between model performance and \textit{Similarity Trajectory}. The first experiment examines how the averaged \textit{Similarity Trajectory} evolves as the training progresses. The second experiment investigated how the average \textit{Similarity Trajectory} changes as the model size increases.

We hypothesize that a better-performed image generation model, characterized by its ability to produce higher quality images, achieves increased similarity scores by generating fewer artifacts that detract from similarity. Hence, as training progresses, models' ability to generate better-quality images improves and hence likely to produce images with fewer artifacts. Additionally, larger models are expected to generate better-quality images in general. Therefore, the similarity between denoised latents will also be higher. 

For the first experiment, we utilize the checkpoints of the large model with a fixed exponential moving average length parameter \(\sigma_{\text{rel}} = 0.05\), as provided by the EDM2 repository \cite{Karras2024edm2}. The first snapshot (T1) occurs after training on 67 million images. The training then continues for an additional 128 million images to produce the second snapshot (T2). Following the same pattern, the third snapshot (T3) is taken after another 128 million images of training, and so forth, ultimately producing seven models (T1 through T7) at different stages of training \cite{Karras2024edm2}.

\begin{figure}[h]
\centering
\begin{subfigure}[t]{\columnwidth}
\centering
\includegraphics[width=\columnwidth]{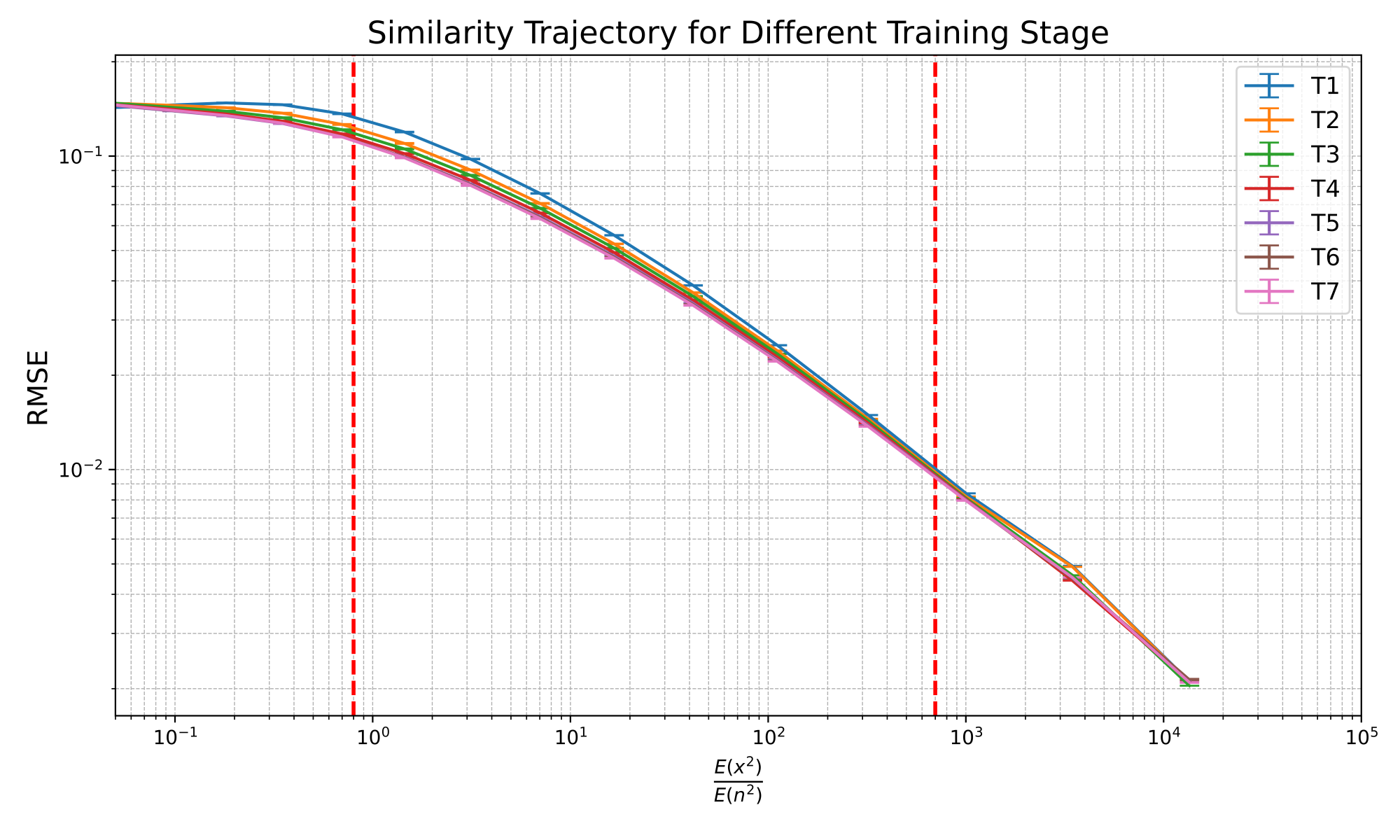}
\caption{Averaged trajectory at different training stages.}
\label{pic:t_stage}
\end{subfigure}

\vspace{0.3cm}

\begin{subfigure}[t]{\columnwidth}
\centering
\includegraphics[width=\columnwidth]{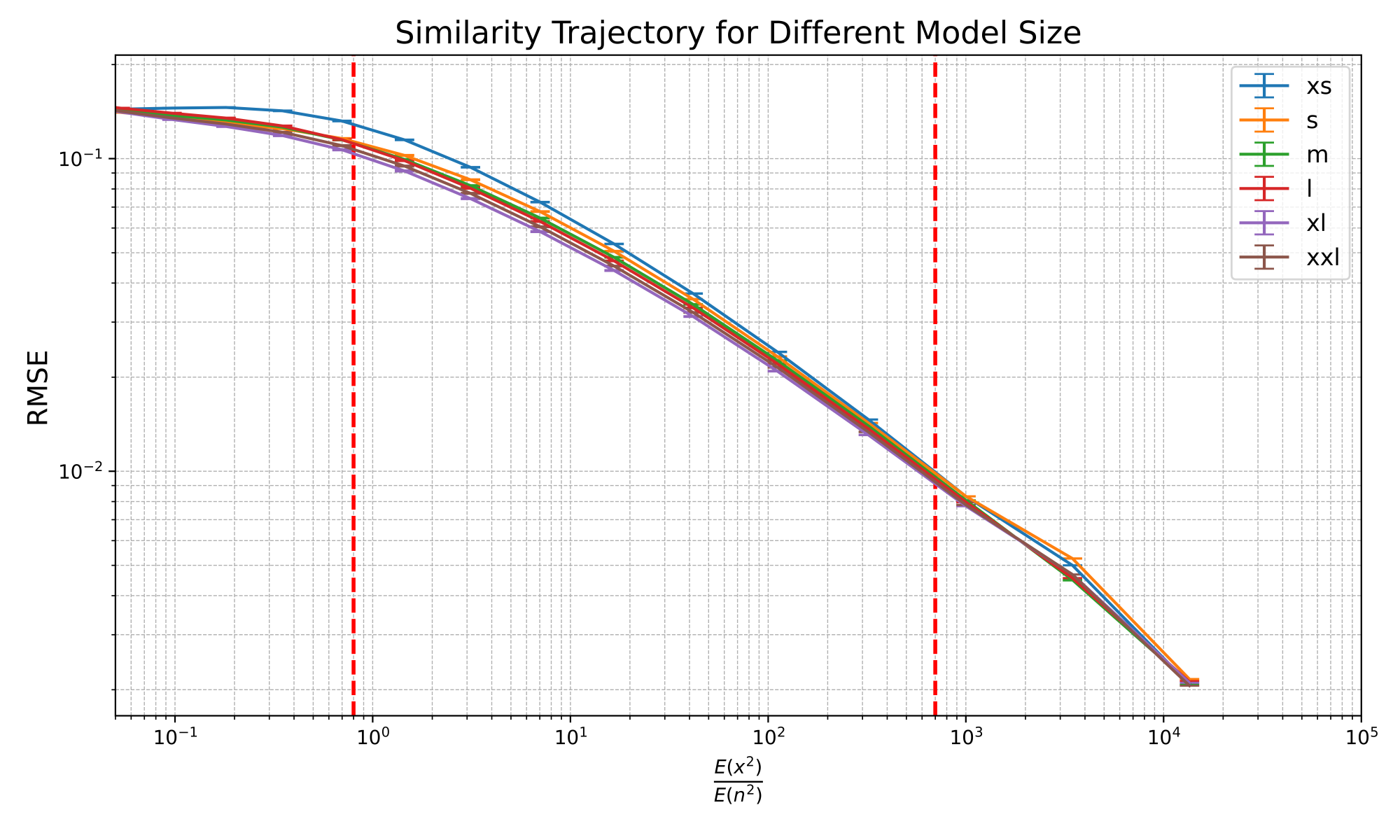}
\caption{Averaged trajectory for fully trained models of varying sizes. }
\label{pic:model_size}
\end{subfigure}

\caption{Comparison of averaged trajectories under different conditions. (a) Effect of training progress on the same model, showing increased latent consistency over time. (b) Influence of model size on fully trained models, with larger models exhibiting greater adjacent latent similarity.}
\label{pic:combined}
\end{figure}

In Figures  \ref{pic:t_stage} and \ref{pic:model_size}, we present the average Root Mean Square Error (RMSE) between denoised latents from consecutive time steps during the sampling process, along with the SEM. The y-axis represents this average RMSE, where larger values indicate more dissimilar latents. The x-axis represents the Signal-to-Noise Ratio (SNR) of the noisy latents, calculated based on the noise level at each time step during training. The latents are normalized to have a mean of 0 and a standard deviation of 0.5.

We chose to use SNR for the x-axis instead of time steps because the time step arrangement differs between the SD2 and EDM frameworks. Reporting SNR directly avoids confusion and provides a clearer assessment of the noise level. We display only the portion of the averaged trajectory where the noisy latents' SNR ranges from \(5 \times 10^{-2}\) to \(10^4\). We excluded SNR values below \(5 \times 10^{-2}\) because, at such low levels, the sampling process cannot extract meaningful information from the noisy latents. This scenario falls outside the scope of our discussion.

Within the region marked by the two red lines (SNR between \(8 \times 10^{-1}\) and \(7 \times 10^{2}\)) in Figure~\ref{pic:t_stage}, a clear trend emerges: as training progresses, the RMSE score between adjacent denoised latents gradually decreases compared to earlier trained models. This matches our initial assumption that as the model becomes stronger, a higher average similarity is expected between adjacent denoised latents. As training progresses, we notice an increasing similarity between denoised latents at adjacent time steps, but the differences in RMSE scores across different training stages are small. However, the results are statistically significant, as the SEM is even smaller than the gap between trajectories, as shown in the graph. 

In Figure~\ref{pic:model_size}, we observe a similar trend: as model capacity increases, the similarity between denoised latents at adjacent time steps also increases. The only exception is a change in the order between the XL and XXL models, which we attribute to the possibility that the XXL and XL models may have similar performance.

Overall, we demonstrate empirically that a correlation between the similarity of adjacent denoised latents and both training time and model capacity exists. This suggests that stronger models—whether due to extended training or increased capacity—are associated with higher similarity scores. This phenomenon could potentially lead to an additional performance indicator, possibly complementing the existing FID score. However, these findings are still preliminary. To determine whether the averaged similarity score is truly a reliable indicator, further evaluations across different models would be necessary.

\section{Conclusion and Limitations}
In this paper, we introduce the concept of the \textit{Similarity Trajectory} and demonstrate its relationship with the presence of artifacts in images generated by diffusion models. Our findings suggest that even with limited training data, 680 labeled images, a classifier can still utilize \textit{Similarity Trajectory} to select artifact-exhibiting images. 

Although the method requires only sparse training data, our work has several limitations. The classification accuracy of the RF Classifier based on the \textit{Similarity Trajectory} is not perfect. Our results suggest that while inconsistencies in the \textit{Similarity Trajectory} may signal potential image artifacts, artifacts may originate from different sources, complicating their identification. Evaluating artifacts directly from the final image offers a more intuitive approach. However, the wide range of possible image distortions makes this method both challenging and data-intensive. Future work could explore integrating these two methods to capitalize on their complementary strengths, thereby improving detection accuracy and generalizability.

For model evaluation, our results suggest that a more capable model is associated with higher similarity between adjacent latents in the EDM framework. However, it would be intriguing to investigate whether diffusion models trained under different settings exhibit similar behavior. We hope our study sparks further exploration into the connection between the sampling process, image quality, and generator performance.

\clearpage

{\small
\bibliographystyle{ieee_fullname}
\bibliography{egbib}

\begin{thebibliography}{10}\itemsep=-1pt

\bibitem{breiman2001random}
Leo Breiman.
\newblock Random forests.
\newblock {\em Machine Learning}, 45:5--32, 2001.

\bibitem{chung2023diffusion}
Hyungjin Chung, Jeongsol Kim, Michael~Thompson Mccann, Marc~Louis Klasky, and Jong~Chul Ye.
\newblock Diffusion posterior sampling for general noisy inverse problems.
\newblock In {\em The Eleventh International Conference on Learning Representations}, 2023.

\bibitem{Cover67knn}
Thomas~M. Cover and Peter~E. Hart.
\newblock Nearest neighbor pattern classification.
\newblock {\em IEEE Transactions on Information Theory}, 13:21--27, 1967.

\bibitem{nips2021_dmbeatsgan}
Prafulla Dhariwal and Alexander Nichol.
\newblock Diffusion models beat gans on image synthesis.
\newblock In M. Ranzato, A. Beygelzimer, Y. Dauphin, P.S. Liang, and J.~Wortman Vaughan, editors, {\em Advances in Neural Information Processing Systems}, volume~34, pages 8780--8794. Curran Associates, Inc., 2021.

\bibitem{fu2023dreamsim}
Stephanie Fu, Netanel~Yakir Tamir, Shobhita Sundaram, Lucy Chai, Richard Zhang, Tali Dekel, and Phillip Isola.
\newblock Dreamsim: Learning new dimensions of human visual similarity using synthetic data.
\newblock In {\em Thirty-seventh Conference on Neural Information Processing Systems}, 2023.

\bibitem{NIPS2014GAN}
Ian Goodfellow, Jean Pouget-Abadie, Mehdi Mirza, Bing Xu, David Warde-Farley, Sherjil Ozair, Aaron Courville, and Yoshua Bengio.
\newblock Generative adversarial nets.
\newblock In Z. Ghahramani, M. Welling, C. Cortes, N. Lawrence, and K.Q. Weinberger, editors, {\em Advances in Neural Information Processing Systems}, volume~27. Curran Associates, Inc., 2014.

\bibitem{fid_Martin}
Martin Heusel, Hubert Ramsauer, Thomas Unterthiner, Bernhard Nessler, and Sepp Hochreiter.
\newblock Gans trained by a two time-scale update rule converge to a local nash equilibrium.
\newblock In {\em Advances in Neural Information Processing Systems}, 2017.

\bibitem{neurips2020ddpm}
Jonathan Ho, Ajay Jain, and Pieter Abbeel.
\newblock Denoising diffusion probabilistic models.
\newblock In H. Larochelle, M. Ranzato, R. Hadsell, M.F. Balcan, and H. Lin, editors, {\em Advances in Neural Information Processing Systems}, volume~33, pages 6840--6851. Curran Associates, Inc., 2020.

\bibitem{karras2022elucidating}
Tero Karras, Miika Aittala, Timo Aila, and Samuli Laine.
\newblock Elucidating the design space of diffusion-based generative models.
\newblock In {\em Advances in Neural Information Processing Systems}, 2022.

\bibitem{Karras2024edm2}
Tero Karras, Miika Aittala, Jaakko Lehtinen, Janne Hellsten, Timo Aila, and Samuli Laine.
\newblock Analyzing and improving the training dynamics of diffusion models.
\newblock In {\em Proc. CVPR}, 2024.

\bibitem{Kawar_2022_Denoising}
Bahjat Kawar, Michael Elad, Stefano Ermon, and Jiaming Song.
\newblock Denoising diffusion restoration models.
\newblock In {\em Advances in Neural Information Processing Systems}, 2022.

\bibitem{kingma2022vae}
Diederik~P Kingma and Max Welling.
\newblock Auto-encoding variational bayes, 2022.

\bibitem{neurips2023Kirstain}
Yuval Kirstain, Adam Polyak, Uriel Singer, Shahbuland Matiana, Joe Penna, and Omer Levy.
\newblock Pick-a-pic: An open dataset of user preferences for text-to-image generation.
\newblock In A. Oh, T. Naumann, A. Globerson, K. Saenko, M. Hardt, and S. Levine, editors, {\em Advances in Neural Information Processing Systems}, volume~36, pages 36652--36663. Curran Associates, Inc., 2023.

\bibitem{Li2022SRDiff}
Haoying Li, Yifan Yang, Meng Chang, Shiqi Chen, Huajun Feng, Zhihai Xu, Qi Li, and Yueting Chen.
\newblock Srdiff: Single image super-resolution with diffusion probabilistic models.
\newblock {\em Neurocomputing}, 2022.

\bibitem{Liang_2024_rich}
Youwei Liang, Junfeng He, Gang Li, Peizhao Li, Arseniy Klimovskiy, Nicholas Carolan, Jiao Sun, Jordi Pont-Tuset, Sarah Young, Feng Yang, Junjie Ke, Krishnamurthy~Dj Dvijotham, Katherine~M. Collins, Yiwen Luo, Yang Li, Kai~J Kohlhoff, Deepak Ramachandran, and Vidhya Navalpakkam.
\newblock Rich human feedback for text-to-image generation.
\newblock In {\em Proceedings of the IEEE/CVF Conference on Computer Vision and Pattern Recognition (CVPR)}, pages 19401--19411, June 2024.

\bibitem{cocodataset}
Tsung{-}Yi Lin, Michael Maire, Serge~J. Belongie, Lubomir~D. Bourdev, Ross~B. Girshick, James Hays, Pietro Perona, Deva Ramanan, Piotr Doll{'{a} }r, and C.~Lawrence Zitnick.
\newblock Microsoft {COCO:} common objects in context.
\newblock {\em CoRR}, abs/1405.0312, 2014.

\bibitem{Lugmayr_2022_CVPR}
Andreas Lugmayr, Martin Danelljan, Andres Romero, Fisher Yu, Radu Timofte, and Luc Van~Gool.
\newblock Repaint: Inpainting using denoising diffusion probabilistic models.
\newblock In {\em Proceedings of the IEEE/CVF Conference on Computer Vision and Pattern Recognition (CVPR)}, pages 11461--11471, June 2022.

\bibitem{podell2024sdxl}
Dustin Podell, Zion English, Kyle Lacey, Andreas Blattmann, Tim Dockhorn, Jonas M{\"u}ller, Joe Penna, and Robin Rombach.
\newblock {SDXL}: Improving latent diffusion models for high-resolution image synthesis.
\newblock In {\em The Twelfth International Conference on Learning Representations}, 2024.

\bibitem{pmlr-v139-radford21a}
Alec Radford, Jong~Wook Kim, Chris Hallacy, Aditya Ramesh, Gabriel Goh, et~al.
\newblock Learning transferable visual models from natural language supervision.
\newblock In {\em Proceedings of the 38th International Conference on Machine Learning}, 2021.

\bibitem{Rombach_2022_ldm}
Robin Rombach, Andreas Blattmann, Dominik Lorenz, Patrick Esser, and Bj\"orn Ommer.
\newblock High-resolution image synthesis with latent diffusion models.
\newblock In {\em Proceedings of the IEEE/CVF Conference on Computer Vision and Pattern Recognition (CVPR)}, June 2022.

\bibitem{unet}
Olaf Ronneberger, Philipp Fischer, and Thomas Brox.
\newblock U-net: Convolutional networks for biomedical image segmentation.
\newblock In {\em Medical Image Computing and Computer-Assisted Intervention - {MICCAI} 2015 - 18th International Conference Munich, Germany, October 5 - 9, 2015, Proceedings, Part {III}}, volume 9351 of {\em Lecture Notes in Computer Science}, pages 234--241. Springer, 2015.

\bibitem{song2021denoising}
Jiaming Song, Chenlin Meng, and Stefano Ermon.
\newblock Denoising diffusion implicit models.
\newblock In {\em International Conference on Learning Representations}, 2021.

\bibitem{stankovics200325}
Radomir~S. Stanković and Bogdan~J. Falkowski.
\newblock The haar wavelet transform: its status and achievements.
\newblock {\em Computers and Electrical Engineering}, 29(1):25--44, 2003.

\bibitem{wu2023hps}
Xiaoshi Wu, Keqiang Sun, Feng Zhu, Rui Zhao, and Hongsheng Li.
\newblock Better aligning text-to-image models with human preference.
\newblock {\em ArXiv}, abs/2303.14420, 2023.

\bibitem{xu2023imagereward}
Jiazheng Xu, Xiao Liu, Yuchen Wu, Yuxuan Tong, Qinkai Li, Ming Ding, Jie Tang, and Yuxiao Dong.
\newblock Imagereward: Learning and evaluating human preferences for text-to-image generation.
\newblock In {\em Thirty-seventh Conference on Neural Information Processing Systems}, 2023.

\bibitem{Zhang_2023_ICCV}
Lingzhi Zhang, Zhengjie Xu, Connelly Barnes, Yuqian Zhou, Qing Liu, He Zhang, Sohrab Amirghodsi, Zhe Lin, Eli Shechtman, and Jianbo Shi.
\newblock Perceptual artifacts localization for image synthesis tasks.
\newblock In {\em Proceedings of the IEEE/CVF International Conference on Computer Vision (ICCV)}, pages 7579--7590, October 2023.

\bibitem{zhang2023shiftddpms}
Zijian Zhang, Zhou Zhao, Jun Yu, and Qi Tian.
\newblock Shiftddpms: Exploring conditional diffusion models by shifting diffusion trajectories.
\newblock {\em arXiv preprint arXiv:2302.02373}, 2023.

\end{thebibliography}
}

\clearpage

\section{Supplementary Material}
\subsection*{Detailed Information for the Experimental Setup}
We define the \textit{Similarity Trajectory} as a discrete time series $\{ z_t \}_{t=T-1}^{1}$, where $T$ is the total number of time steps in the sampling process. Each element $z_t$ represents the similarity score between the denoised images in consecutive time steps $t$ and $t-1$. To analyze fluctuations within this trajectory, we segment it either based on time steps or by projecting it onto different bases using the Haar Transform. We extract various statistical quantities to characterize each set which subsequently serve as inputs to a RF Classifier. Segmenting the Similarity Trajectory in the time domain allows us to capture variations at specific time steps, which convey varying information; notably, fluctuations during the middle section of sampling are critical for artifact detection. Additionally, the coefficients from different levels of the Haar Transform reveal fluctuations at various frequencies, enabling us to characterize the duration and magnitude of these changes. 

\subsection*{Segmentation of the Time Series}
We divide the entire \textit{Similarity Trajectory} into three equal sets $S_1, S_2, S_3$ based on time steps, as well as considering the entire series as a single set, denoted as $S_4$. Segments in the time domain are formally defined as:
\begin{itemize}
    \item \textbf{Segment 1 ($S_1$)}:
    \[
    S_1 = \{ z_t \mid t = 1, 2, \dotsc, N_1 \},
    \]
    where
    \[
    N_1 = \left\lfloor \dfrac{T - 1}{3} \right\rfloor.
    \]
    \item \textbf{Segment 2 ($S_2$)}:
    \[
    S_2 = \{ z_t \mid t = N_1 + 1, N_1 + 2, \dotsc, N_2 \},
    \]
    where
    \[
    N_2 = \left\lfloor \dfrac{2(T - 1)}{3} \right\rfloor.
    \]
    \item \textbf{Segment 3 ($S_3$)}:
    \[
    S_3 = \{ z_t \mid t = N_2 + 1, N_2 + 2, \dotsc, T - 1 \}.
    \]
    \item \textbf{Segment 4 ($S_4$)}:
    \[
    S_4 = \{ z_t \mid t = 1, 2, \dotsc, T - 1 \}.
    \]
\end{itemize}

\subsection*{Segmentation of Coefficients for Haar Transform}
We apply the discrete Haar wavelet transform to the entire \textit{Similarity Trajectories} $\{ z_t \}_{t=T-1}^{1}$ because of its ability to detect sudden changes in time-series data \cite{stankovics200325}. In this section, we first introduce the Haar Transform, and then we explain how to process the \textit{Similarity Trajectory} using the Haar Transform.

\subsubsection*{Haar Transformation}
The Haar Transform decomposes the original time series into approximation and detail coefficients at various scales, capturing both global trends and local variations.

At the first level of decomposition, for $k = 1, 2, \dots, \left\lfloor \dfrac{T-1}{2} \right\rfloor$, the approximation coefficients $a_1(k)$ and detail coefficients $d_1(k)$ are calculated as:
\begin{equation}
a_1(k) = \frac{ z_{2k-1} + z_{2k} }{ 2},
\end{equation}
\begin{equation}
d_1(k) = \frac{ z_{2k-1} - z_{2k} }{ 2 }.
\end{equation}

This process is recursively applied to the approximation coefficients to obtain higher-level coefficients. At level $j+1$, the coefficients are computed as:
\begin{equation}
a_{j+1}(k) = \frac{ a_j(2k-1) + a_j(2k) }{ 2 },
\end{equation}
\begin{equation}
d_{j+1}(k) = \frac{ a_j(2k-1) - a_j(2k) }{ 2 },
\end{equation}
where $j = 1, 2, \dots, J$, and $J$ is the maximum level of decomposition. From the transformation, we obtain a set of detail coefficients $\{ d_j(k) \}$ corresponding to each basis function at various levels. Note that the detail coefficients capture the fluctuation information of the \textit{Similarity Trajectory}, which is important for assessing the presence of artifacts in the image.

We segment the detail coefficients obtained from the Haar Transform of the entire \textit{Similarity Trajectory} $\{ z_t \}_{t=T-1}^{1}$ by grouping them according to their corresponding Haar basis functions. Each set $S_j$ consists of all detail coefficients at decomposition level $j$:

\[
S_j = \{ d_j(k) \mid k = 1, 2, \dotsc, N_j \},
\]
where $N_j = \left\lceil \dfrac{T - 1}{2^j} \right\rceil$ is the number of coefficients at level $j$. This segmentation aligns each set of detail coefficients with their respective scales in the time series, allowing us to analyze fluctuations captured by each Haar basis function effectively.

\subsection*{Feature Extraction}
For all the sets $S$ obtained—whether from the detail coefficients of the Haar Transform or in the time domain—we calculate ten statistical features for each set to perform the bag-of-statistics method. These statistical features are chosen to describe the dynamics of the \textit{Similarity Trajectory} as we already established that the fluctuation in\textit{ Similarity Trajectory} is correlated to the presence of artifacts. They include:

\begin{enumerate}
    \item \textbf{Mean ($\mu_S$)}: The average value of the data in the set $S$.
    
    \begin{equation}
    \mu_S = \frac{1}{N_S} \sum_{s \in S} s,
    \end{equation}
    where $N_S$ is the number of elements in the set $S$.

    \item \textbf{Standard Deviation ($\sigma_S$)}: Measures the dispersion of the data in the set $S$. This is related to fluctuation in the \textit{Similarity Trajectory} for sets in the time domain.
    \begin{equation}
    \sigma_S = \sqrt{ \frac{1}{N_S } \sum_{s \in S} \left( s - \mu_S \right)^2 }.
    \end{equation}

  \item \textbf{Percentile}: We extract the 5th, 25th, 50th, 75th, and 95th percentiles for each obtained set's values. The significance of percentiles lies in their relation to the fluctuation of the \textit{Similarity Trajectory} for detail coefficients.

    \item \textbf{Number of Mean Crossings ($C_{\mu,S}$)}: Counts how many times the data crosses its mean value in the set $S$. This describes how rapidly the \textit{Similarity Trajectory} fluctuates in the time domain.
    \begin{equation}
    C_{\mu,S} = \sum_{i = 1}^{N_S} \mathbb{I} \left[ \left( S_{i+1} - \mu_s \right) \left( S_i - \mu_s \right) < 0 \right],
    \end{equation}
    where $N_S$ is the total number of elements in set $S$. $S_i$ is the $i^{th}$ element in $S$ and $\mathbb{I}[\cdot]$ is the indicator function.

    \item \textbf{Number of Zero Crossings ($C_{0,S}$)}: Counts how many times the data crosses zero in set $S$. Note that in detail coefficients, this represents how many times the \textit{Similarity Trajectory} changes direction, from monotonically increasing to monotonically decreasing or vice versa.
    \begin{equation}
    C_{0,S} = \sum_{i = 1}^{N_S} \mathbb{I} \left[ S_{i+1} S_i < 0 \right].
    \end{equation}
    where $N_S$ is the total number of elements in set $S$. $S_i$ is the $i^{th}$ element in $S$ and $\mathbb{I}[\cdot]$ is the indicator function.

    \item \textbf{Entropy ($E_s$)}: Measures how uniform of the data in set $S$ is. Again, this is another metric characterizing the fluctuations of the set.
    \begin{equation}
    E_S = -\sum_{i} p_i^{(S)} \log_2 p_i^{(S)},
    \end{equation}
    where $p_i^{(S)}$ is the probability of the $i$-th bin in the histogram of the data in set $S$.

\end{enumerate}

These features are computed for both the time-domain data and the detail coefficients from Haar Transform, resulting in a comprehensive feature set that captures both temporal and frequency-domain characteristics.

\subsection*{Using $k$-Nearest Neighbor Model Probabilities}
We employed a $k$-Nearest Neighbor ($k$-NN) model trained directly on the time-domain \textit{Similarity Trajectory}. This $k$-NN model estimates the probability of artifact presence by assessing the proportion of its nearest neighbors that are labeled as artifact or non-artifact images. These predicted probabilities are then incorporated as additional features into the RF Classifier.

\subsection*{Feature Vector Construction}
For every set $S$, we formulate a feature vector $\mathbf{f}_S$ which includes ten statistical features. The comprehensive feature vector $\mathbf{F}$ for the trajectory is then assembled by concatenating all $\mathbf{f}_S$ vectors from every set along with the prediction probability from the $k$-NN model. The feature vector $\mathbf{F}$ serves as the input to the RF Classifier for both training and inference.

\end{document}